\documentclass[12pt]{article}
\usepackage{amssymb,amsmath,cite,bm}
\usepackage[dvips]{graphicx,color}
\usepackage{psfrag,subfigure}
\begin{document}

\title{\bf Automated Rendezvous \& Docking Using 3D Vision}

\author{Farhad Aghili\thanks{email: faghili@encs.concordia.ca}}

\date{}

\maketitle

\begin{abstract}
The robustness and accuracy of a vision system for motion estimation of a tumbling target satellite are enhanced by an adaptive Kalman filter. This allows a vision-guided robot to complete the grasping of the target even if occlusion occurs during the operation. A complete dynamics model, including aspects of orbital mechanics, is incorporated for accurate estimation. Based on the model, an adaptive Kalman filter is developed that estimates not only the system states but also all the model parameters such as the inertia ratio, center-of-mass, and the rotation of the principal axes of the target satellite. An experiment is conducted by using a robotic arm to move a satellite mockup according to orbital mechanics while the satellite pose is measured by a laser camera system. The measurements are sent to the Kalman filter, which, in turn, drives another robotic arm to grasp the target. The results demonstrate successful grasping even if the vision system is blocked for several seconds.
\end{abstract}

\section{Introduction}
Internationally, there has been growing interest in using space
robotic systems for the on-orbit servicing of
spacecraft~\cite{Aghili-2011k,Bornschlegl-Hirzinger-Maurette-Mugunolo-Visentin-2003,Aghili-2012b,Aghili-Parsa-2007b,
yoshida-2003,Aghili-Parsa-2009b,Whelan-Adler-Wilson-Roesler-2000,Aghili-2016c,Aghili-Parsa-2008b}.  In
such missions, the accurate motion estimation of a free-falling
target spacecraft is essential for guiding a robotic arm so as to
capture the target. Different vision systems are available for the
estimation of the pose (position and orientation) of moving objects.
Among them, an active vision system such as the Neptec Laser Camera
system (LCS) is preferable for its robustness in face of the harsh
lighting conditions of
space~\cite{Samson-English-Deslauriers-Christie-2004}. As
successfully verified during the STS-105 space mission, the 3D
imaging technology used in LCS can indeed operate in space
environment. The use of laser range data has also been proposed for
the motion estimation of free-floating space
objects~\cite{Hillenbrand-Lampariello-2005,Aghili-Kuryllo-Okouneva-English-2010a,Aghili-Parsa-2009,Aghili-2008c,Aghili-Parsa-Martin-2008a,Aghili-2010n}. All vision systems,
however, provide discrete and noisy pose data at relatively low
rate, typically 1~Hz, while the capture of a free-falling object is
a difficult robotic task which requires real-time and accurate pose
estimation.

Taking advantage of the simple dynamics of a free-floating object,
researchers have employed different observers to track and predict
the motion of a target
satellite~\cite{Masutani-Iwatsu-Miyazaki-1994,
Hillenbrand-Lampariello-2005}.  In some circumstances, e.g., when
there are occlusions, no observation data is available. Therefore,
long-term prediction of the motion of the object is needed for
planning operations such as autonomous grasping of
targets~\cite{Masutani-Iwatsu-Miyazaki-1994,
Hillenbrand-Lampariello-2005}.

The Kalman filter uses a dynamics model to compute a rough estimate
of the system states, which is then corrected using a model of the
sensor measurements to obtain the best estimate possible of the
system states based on the present and past measurements. However,
the applicability of the Kalman filtering technique rests on the
accuracy of the dynamics and the measurement-noise models.

In this work, we use the Euler-Hill equations to derive a
discrete-time model that captures the evolution of the relative
translational motions of a tumbling target satellite with respect to
a chaser satellite which is free-falling in a nearby orbit \cite{Aghili-2010f,Aghili-Parsa-2008b}. Taking
advantage of the structure of the model, we derive numerically
efficient expressions for the state transition matrix and the
covariance of the process noise, both of which are necessary for the
Kalman filter. We also derive the sensitivity matrix of the
observation and show that the observations of the translational and
rotational displacements are coupled. Moreover, the kinematic
properties of the unit-norm quaternions used to represent the
relative orientation of the target satellite, is employed to derive
the associated measurement-noise covariance. The new model allows
{\em additive} quaternion noise while preserving the unit-norm
property of the quaternion. This Kalman filter has been implemented
in Matlab/Simulink and used in real-time to permit the autonomous
capture of a tumbling satellite using a chaser arm. The performance
of the Kalman filter is such that it is possible to fully occlude
the vision system and perform the capture of the satellite more than
20 seconds after the occlusion using only the output of the EKF to
provide a prediction of the pose of the satellite.

\section{Modelling} \label{sec:modeling}
\subsection{Linearization}

\begin{figure}
\psfrag{Target_satellite}[c][c][.8]{Target Satellite}
\psfrag{Chaiser_satellite}[c][c][.8]{Chaser Satellite}
\psfrag{Handel}[c][c][.7]{Handle} \psfrag{Camera}[c][c][.7]{Camera}
\psfrag{por}[c][c][.7]{POR} \psfrag{FA}[c][c][.8]{$\{A \}$}
\psfrag{FB}[c][c][.8]{$\{ B\}$} \psfrag{FC}[c][c][.8]{$\{ C\}$}
\psfrag{rm}[c][c][.8]{$r_m$} \psfrag{omeg}[c][c][.8]{$\omega $}
\psfrag{ro}[c][c][.8]{$r_o$} \psfrag{rc}[c][c][.8]{$r_c$}
\psfrag{rho_t}[c][c][.8]{$\rho_t$}
\psfrag{rho_c}[c][c][.8]{$\rho_c$}
\centering{\includegraphics[clip,width=10cm]{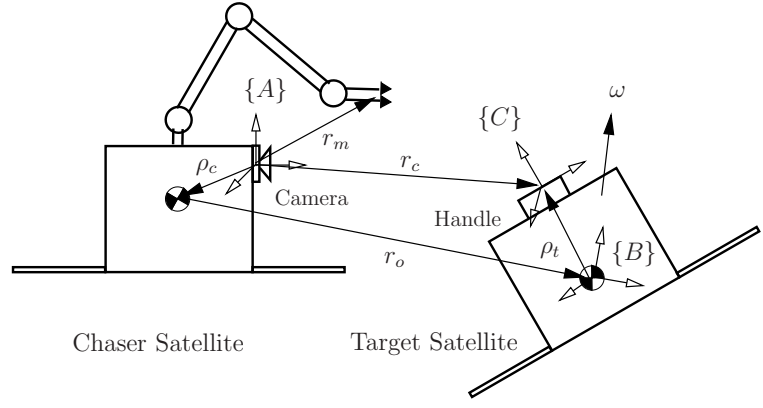}}
\caption{The body-diagram of chaser and target satellites}
\label{fig:chaisersat}
\end{figure}

Figure~\ref{fig:chaisersat} illustrates the chaser and the target
satellites as rigid bodies moving in orbits nearby each other.
Coordinate frames $\{ A \}$ and $\{ B\}$ are attached to the chaser
and the target, respectively. The origin of $\{ B \}$ is located at
the target centre of mass (CM) while that of the camera frame $\{ A
\}$ has an offset $\rho_c$ with respect to the CM of the chaser. The
axes of $\{ B \}$ are oriented so as to be parallel to the principal
axes of the target satellite. Coordinate frame $\{ C \}$ is fixed to
the target at its POR located at $\rho_t$ from the origin of $\{ B
\}$; it is the pose of $\{ C \}$ which is measured by the laser
camera. We further assume that the target satellite tumbles with
angular velocity $\omega$. Also, notice that coordinate frame $\{ A
\}$ is not inertial; rather, it moves with the chaser satellite. In
the following, quantities $\rho_t$ and $\omega$ are expressed in $\{
B \}$, while $\rho_c$ is expressed in $\{ A \}$.

The orientation of $\{ B\}$ with respect to $\{ A\}$ is represented
by the unit quaternion $q$. Below, we review some basic definitions
and properties of quaternions used in the rest of the paper.
Consider quaternion $q_1$, $q_2$, $q_3$, and their corresponding
rotation matrix $R_1$, $R_2$, and $R_3$. The operators $\otimes$ and
$\odot$ are defined as
\[ [a \otimes ] \triangleq \begin{bmatrix}  - \lceil a_v \rceil + a_o I_3 & a_v\\-a_v^T&
a_o \end{bmatrix}, \quad [a \odot ] \triangleq
\begin{bmatrix}  \lceil a_v \rceil + a_o I_3 & a_v\\ - a_v^T  & a_o
\end{bmatrix},
\]
where $a_o$ and $a_v$ are the scalar and vector parts of quaternion
$a$, respectively, and $\lceil a_v \rceil$ is the cross-product
matrix of $a_v$. Then,
\[ q_3 = q_1 \otimes q_2 = q_2 \odot q_1, \]
corresponds to product $R_3=R_1 R_2$.

Consider a small quaternion perturbation
\begin{equation}\label{eq:deltaq-def}
\delta q=q\otimes\bar q^*
\end{equation}
where $q$ represents the rotation of the target satellite with
respect to the chaser satellite. Then, one can linearize the above
equation about the estimated states $\bar q$ and $\bar\omega$ to
obtain \cite{Aghili-Salerno-2016}
\begin{equation} \label{eq:delta_qv}
\frac{d}{dt} \delta q_v \approx - \bar \omega_k \times \delta q_v +
\frac{1}{2} \delta \omega
\end{equation}

Dynamics of the rotational motion of the target satellite can be
expressed by Euler's equation as
\begin{equation} \label{eq:dot_omega}
\dot \omega = \psi(\omega) + J \varepsilon_{\tau}, \quad
\text{where} \quad
\psi(\omega) = \begin{bmatrix} p_x \omega_y \omega_z \\ p_y \omega_x \omega_z \\
p_z \omega_x \omega_y  \end{bmatrix},
\end{equation}
where $J=\text{diag}\big(1, {I_{xx}}/{I_{yy}}, {I_{xx}}/{I_{zz}}
\big)$; $p_x=(I_{yy}- I_{zz})/I_{xx}$, $p_y=(I_{zz}-
I_{xx})/I_{yy}$, and $p_z=(I_{xx}- I_{yy})/I_{zz}$; $I_{xx}$,
$I_{yy}$, and $I_{zz}$ are the principal moments of inertia of the
target satellite; $\varepsilon_{\tau}$ is a torque disturbance for
unit inertia. Defining $p=\begin{bmatrix}p_x & p_y & p_z
\end{bmatrix}$ and linearizing~\eqref{eq:dot_omega} about $\bar
\omega_k$  and $\bar p_k$ yields
\begin{equation} \label{eq:dot_omeg_param_lin}
\frac{d}{d t} \delta \omega = A(\bar \omega_k, \bar p) \delta \omega
+ B(\bar \omega_k) \delta p + J \varepsilon_{\tau},
\end{equation}
where \begin{equation*}
 A(\omega,p) = \frac{\partial\psi}{\partial\omega} =
\begin{bmatrix} 0 & p_x \omega_{z} & p_x \omega_{y} \\ p_y \omega_{z} & 0 &
  p_y \omega_{x} \\ p_z \omega_{y} & p_z \omega_{x} & 0
  \end{bmatrix} \quad \mbox{and} \quad
B(\omega) = \frac{\partial \psi}{\partial p} = \mbox{diag}
\begin{bmatrix} \omega_y \omega_z & \omega_x \omega_z & \omega_x
\omega_y \end{bmatrix}.
\end{equation*}
Let $x=[q_v^T\thickspace\omega^T\thickspace p^T ]^T$ describe the
part of the system states pertaining to the rotational motion. Then,
from \eqref{eq:delta_qv} and \eqref{eq:dot_omeg_param_lin}, we have
\begin{equation} \label{eq:rotation_lin}
\frac{\text d}{{\text d}t} \delta x =
\begin{bmatrix} -\lceil \bar \omega_k  \rceil & \frac{1}{2} I_3 & 0_{3 \times 3}\\
0_{3 \times 3} & A(\bar \omega_k, \bar p_k) & B(\bar \omega_k) \\
0_{3 \times 3} & 0_{3 \times 3} &  0_{3 \times 3}
\end{bmatrix} \delta x + \begin{bmatrix} 0_{3 \times 1} \\ J \epsilon_{\tau} \\ 0_{3 \times 1} \end{bmatrix}.
\end{equation}

In addition to the inertia of target satellite, the location of its
CM and the orientation of the principal axes $\eta_v$ are uncertain.
Let vector $\theta^T =[ \rho_t^T \thickspace \eta_v^T]$ contains the
additional unknown parameters. Then
\begin{equation} \label{eq:dtheta}
\dot \theta = 0
\end{equation}

The evolution of the relative distance of the two satellites can be
described by {\em orbital mechanics}. Let the chaser move on a
circular orbit at an angular rate of $n$ defined as $n=\begin{bmatrix} 0 & 0 & n_z
\end{bmatrix}^T$. Further, assume that
vector $r_o$ denotes the distance between the CMs of the two
satellites expressed in $\{ A\}$, and that $\upsilon_o=\dot r_o$.
Then, if $\{ A\}$ is orientated so that its $x$-axis is radial and
pointing outward, and its $y$-axis lies on the orbital plane, the
translational motion of the target can be expressed as
\cite{Breakwell-Roberson-1970}
\begin{equation} \label{eq:dot_vo}
\dot \upsilon_o = -2 n \times \upsilon_o + \eta(r_o,n) + \epsilon_f.
\end{equation}
Here,  $\epsilon_f$ is the force disturbance for a unit mass, and
acceleration term $\eta$ is due to the effect of orbital mechanics
and can be linearized as $ \eta^T(r_o,n) \approx [ 3n_z^2 r_{o_x} \;
0 \; -n_z^2r_{o_z} ]$. Denoting the states of the translational
motion with $y^T=\begin{bmatrix} r_o^T &  \upsilon_o^T
\end{bmatrix}$, one can derive the corresponding dynamics as
\begin{equation} \label{eq:translation_lin}
\frac{\text d}{{\text d}t} \delta y  =
\begin{bmatrix} 0_{3 \times 3} & I_3 \\
N & -2 \lceil n \rceil
\end{bmatrix} \delta y + \begin{bmatrix} 0_{3 \times 1}  \\ \epsilon_{f} \end{bmatrix}
\quad \mbox{where} \quad
N \triangleq  \frac{\partial \eta}{\partial r_o} = \begin{bmatrix}
3n_z^2 & 0 & 0 \\ 0 & 0 & 0 \\ 0 & 0 & -n_z^2 \end{bmatrix}.
\end{equation}

\subsection{Discrete Model} \label{sec:discrete}

In order to take into account the composition rule of quaternion,
the states to be estimated by the Kalman filter have to be redefined
as $x_k^T=[\delta_k^T~\thickspace\omega_k^T\thickspace~p_k^T]$,
$y_k^T=[r_{o_k}^T\thickspace~y_{o_k}^T]$, and
$\theta_k^T=[\rho_{t_k}^T\thickspace\delta\eta_{v_k}^T]$, where
$\delta \eta_=\bar \eta^* \otimes \eta$. Assuming $
\chi^T\triangleq[x^T\thickspace~y^T\thickspace~\theta^T]$, one can
combine the nonlinear equations \eqref{eq:dot_omega},
\eqref{eq:dtheta} and \eqref{eq:dot_vo} in the standard form as
$\dot \chi = f( \chi, \epsilon)$, where $\epsilon^T=[
\epsilon_{\tau}^T \; \epsilon_f^T ]$. Moreover, setting the
linearized systems \eqref{eq:rotation_lin}, \eqref{eq:dtheta} and
\eqref{eq:translation_lin} in the standard state-space form $\dot
\chi = \mathcal{A} \chi + \mathcal{B} \epsilon$, the equivalent
discrete-time system is
\begin{equation} \label{eq:discere_sys}
\chi_{k+1} = \Phi_k \chi_k + \epsilon_k.
\end{equation}
Where the solution to the state transition matrix $\Phi_k$ and
discrete-time process noise $Q_k=E[\epsilon_k \epsilon_k^T]$ can be
obtained based on the van Loan method  as $\Phi_k=D_{22}^T$ and
$Q_k=\Phi_k D_{12}$, where
\[ D = \begin{bmatrix} D_{11} & D_{12} \\ 0 & D_{22} \end{bmatrix} = \exp \Big( \begin{bmatrix} -\mathcal{A} &
\mathcal{B}\Sigma_{\epsilon} \mathcal{B}^T \\ 0 & \mathcal{A}^T
\end{bmatrix} T \Big)\] with $T$ being the sampling time and
$\Sigma_{\epsilon} = E[ \epsilon
\epsilon^T]=\mbox{diag}(\sigma_{\tau}^2 I_3, \sigma_{f}^2I_3)$.

\section{Observation}
\subsection{Sensitivity Matrix and Propagation of Measurement Noise}

The vision system output contain the pose of frame $\{ C\}$ w.r.t
frame $\{ A\}$, i.e., the position vector $r_c$ and the orientation
$\mu=\eta \otimes q$. Therefore, the observation vector is defined
as
\begin{equation}\label{eq:delta-meas}
z = h(x) +\begin{bmatrix}v\end{bmatrix},
\end{equation}
where $z^T=[z_1^T \thickspace z_2^T]$, $h^T=[h_1^T \thickspace
h_2^T]$, $v^T=[v_1^T \thickspace v_2^T]$, and $v_1$ and $v_2$ are
additive measurement noise processes, and
\begin{align}\label{eq:h1}
h_1 &\triangleq\ r_c - \bar r_c = \delta r_o + R(\bar q)
\big(R(\delta q) -I_3 \big) \rho_t \approx  \delta r_o + 2R(\bar q)
\lceil \delta q_v \rceil \rho_t \\ \label{eq:h2} h_2 &\triangleq
\big(\bar \eta^* \otimes \mu \otimes \bar q^* \big)_v = \big(\delta
\eta \otimes \delta q \big)_v = \big( \delta q \odot \delta \eta
\big)_v .
\end{align}
Then, in view of \eqref{eq:h1} and the following partial derivatives
\[ \frac{\partial h_2}{\partial \delta q_v} = \lceil \delta \eta_v
\rceil + \delta \eta_o I_3 + \delta q_o^{-1} \delta \eta_v \delta
q_v^T, \qquad \frac{\partial h_2}{\partial \delta \eta_v} = -\lceil
\delta q_v \rceil + \delta q_o I_3 + \delta \eta_o^{-1} \delta q_v
\delta \eta_v^T, \] and neglecting the small terms, i.e., $\delta
\eta_v \delta q_v^T \approx 0$, the sensitivity matrix is obtained
as
\begin{equation} \label{eq:Ha}
H_k = \begin{bmatrix}-2R(\bar q)\lceil \rho_{t_k} \rceil  & 0_{3
\times 6} &  I_3 & 0_{3\times3} & R(\bar q) & 0_{3\times 3} \\
\lceil \delta \eta_{v_k} \rceil + \delta \eta_{o_k} I_3    & & 0_{3
\times 15}  & & & -\lceil \delta q_{v_k} \rceil + \delta q_{o_k}
I_3\end{bmatrix}.
\end{equation}
Here we assume that $\delta \eta_v$ is sufficiently small so that
$\delta \eta_o$ can be unequivocally obtained from $\delta \eta_o =
(1 - \| \delta \eta_v \|)^{1/2}$.

Moreover, as shown in \cite{Aghili-Parsa-2007b,Aghili-2005}, in the case of
isotropic orientation noise, i.e., $\Sigma_{q_v} =
\sigma_{q_o}^2I_3$, the equation of the covariance is drastically
reduced to $E[v_2 v_2^T] = \sigma_{q_o}^2 I_3$.

\subsection{Filter design}

Recall that $\delta q_v$ is a small deviation from the the nominal
trajectory $\bar q$. Since the nominal angular velocity $\bar
\omega_k$ is assumed constant during each interval, the trajectory
of the nominal quaternion can be obtained from
\[ \bar q(t) = e^{\frac{1}{2}(t-t_0) [\bar{\underline \omega}_k \otimes]} \bar q(t_0) \quad \Longrightarrow \quad  \bar q_k = e^{\frac{1}{2} T [\hat{\underline \omega}_{k-1} \otimes]} \hat q_{k-1}. \]
Similarly, we have $\bar \eta_k = \hat \eta_{k-1}$. The EKF-based
observer for the associated noisy discrete system
\eqref{eq:discere_sys} is given in two steps: ($i$) estimate
correction
\begin{subequations}
\begin{align} \label{eq:K_est}
K_k & =P_k^-H_k^T \big(H_kP_k^- H_k^T+ S_k \big)^{-1} \\
\label{eq:x_est} \hat \chi_k &= \hat \chi_k^- + K \big(z_k -h(\hat
\chi_k^-) \big) \\ \label{eq:P_est} P_k &= \big( I - K_k H_k \big)
P_k^-
\end{align}
\end{subequations}
and ($ii$) estimate propagation
\begin{subequations}
\begin{align}\label{eq:state-prop}
\hat \chi_{k+1}^- &=\hat \chi_k + \int_{t_k}^{t_{k+1}} f(\chi(t),0)\,{\text d}t\\
P_{k+1}^- &= \Phi_k P_k \Phi_k^T + Q_k
\end{align}
\end{subequations}
and the quaternions are computed right after the innovation step
\eqref{eq:x_est} from
\begin{equation*} \hat q_k = \delta \hat q_k
\otimes \bar q_k \quad \text{and} \quad \hat \eta_k = \delta \hat
\eta_k \odot \bar \eta_k .
\end{equation*}

\section{Experiment}
\begin{figure}[t]
\centering{\includegraphics[clip,width=10cm]{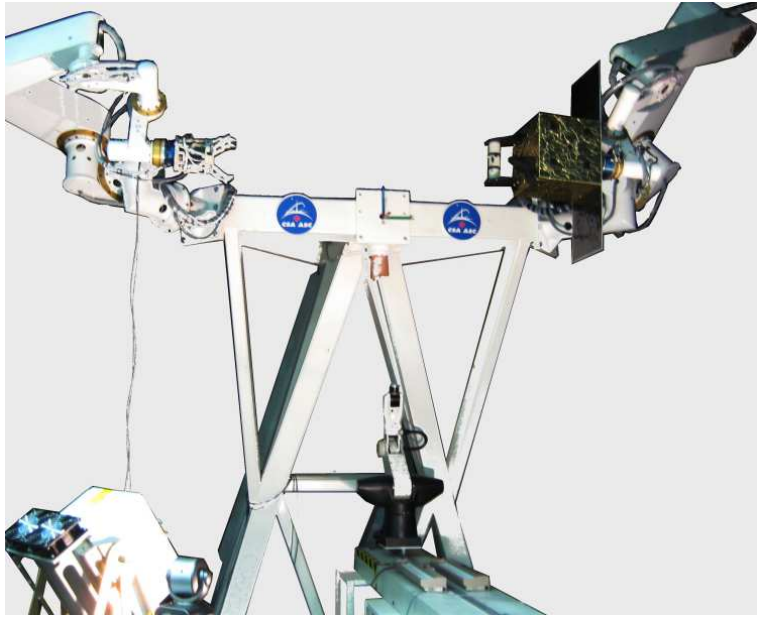}}
\caption{The experimental setup.} \label{fig:lcs_cart}
\end{figure}

\begin{figure}
\centering{\includegraphics[clip,width=12cm]{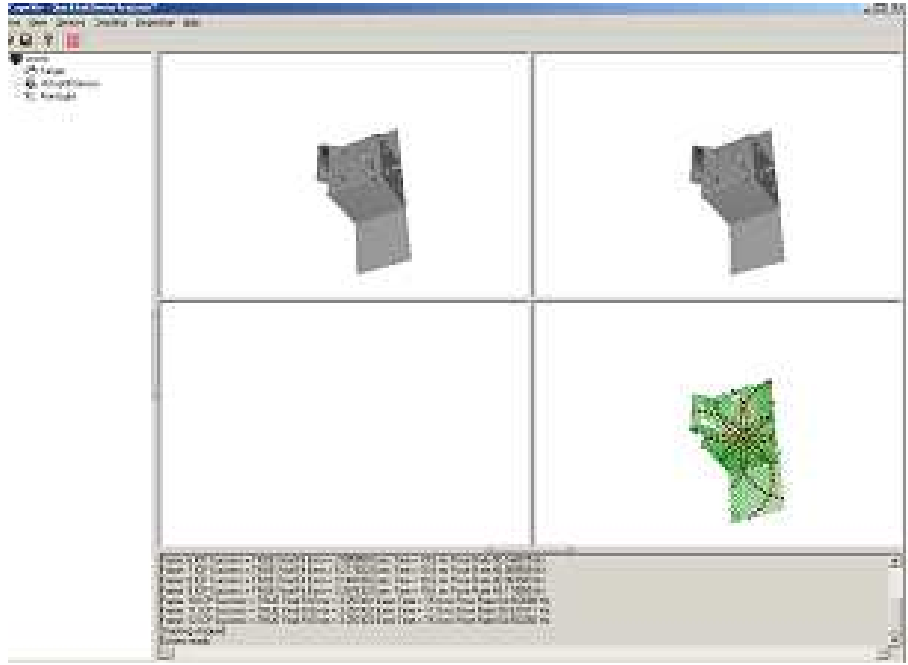}}
\caption{Surface model of target} \label{fig:surface_model}
\end{figure}

In this section, experimental results are presented and show the
performance of the predictor that permits the capture of a tumbling
satellite even if the vision system is fully occluded by providing
the prediction of the pose of the satellite based on the estimate of
the full states and parameters at a given time. The Neptec Laser
Camera system (LCS)
\cite{Samson-English-Deslauriers-Christie-2004}, is used to obtain
the pose measurements at a rate of 2~Hz. Figure~\ref{fig:lcs_cart}
illustrates the experimental setup, whereby the mockup of a
satellite is moved by a manipulator according to orbital dynamics
using the concept of hybrid simulator  \cite{Piedboeuf-Aghili-Doyon-Martin-2002, Aghili-Piedboeuf-2002,Piedboeuf-deCarufel-Aghili-Dupuis-1999,Aghili-Namvar-2008,Aghili-Namvar-Vukovich-2006} 
while another arm, equipped with a robotic hand, is used to autonomously approach and capture this mockup. 
The CAD surface model of target is imported to the 3D registration algorithm for pose estimation, see Fig.~\ref  {fig:surface_model}. For the spacecraft
simulator that drives the manipulator, parameters are selected as
$I=\mbox{diag}
\begin{bmatrix}4 & 8 &
  5 \end{bmatrix}~\mbox{kgm}^2$ and $\rho_t=\begin{bmatrix} -0.15 & 0 & 0
\end{bmatrix}^T~\mbox{m}$.

\begin{figure}
\centering{\includegraphics[clip,width=11cm]{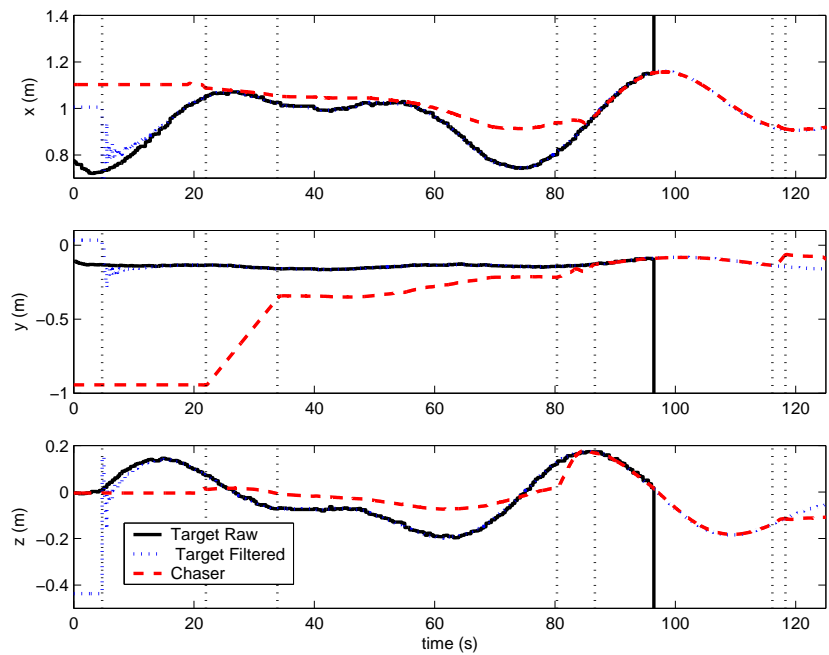}}
\caption{Cartesian position of the grasping device matching the
motion of the target} \label{fig:traj_capt_exp}
\end{figure}
The objective of this experiment is that the robotic arm
autonomously capture the handle of the target satellite using the
pose provided by the LCS. During the tracking of the handle, the
field of view of the LCS was voluntary fully occluded and hence the
EKF is to reliably provide the pose information.
Figure~\ref{fig:traj_capt_exp} illustrates trajectories of the
position measurements from the LCS (solid lines) versus the EKF
outputs (dotted lines) and those of robot end-effector (dashed
lines). The EKF was enabled at t=5~sec. and converges after about
9~sec. later. At t=22~sec., the chaser arm starts approaching the
target satellite and it reaches a safe distance at t=34~sec. This
safe distance is maintained until the autonomy engine determine that
the final alignment can be performed safely. This alignment is
initiated at t=80~sec. and is completed at t=87~sec. At that time,
the chaser arm is tracking the target satellite. At t=96~sec., the
vision system is voluntary fully occluded as we can observe by
looking at the solid black line of Fig.~\ref{fig:traj_capt_exp}.
Even after the failure of the vision system to provide the pose of
the satellite, the EKF continues to provide a good prediction of its
pose and the tracking of the target satellite by the chasing arm
continue for about 20~sec. At t=116~sec., a capture command is
issued and the capture is completed successfully at t=118~sec., thus
showing the excellent performance of the EKF at providing a reliable
pose of the target satellite even after the failure of the vision
system.

\section{Conclusion}
A discrete-time adaptive estimator was developed for estimating the
relative pose of two free-falling satellites that move in close
orbits near each other using position and orientation data provided
by a laser vision system. The
covariance matrix of the measurement noise was modelled in a
state-dependant form. This model allows additive quaternion noise,
while preserving the unit-norm property of the quaternion. The
effects of orbital mechanics was incorporated in the model-based
estimator for both the accurate estimation and the prediction of the
relative motion.  The discrete-time model, including the
state-transition matrix and the covariance of the process noise were
derived in closed form suitable for implementing the extended Kalman
filter in real time. Experimental results illustrating the autonomous capture
of a tumbling satellite with an occluded vision system were reported.
The results demonstrated that the filter could
accurately estimate and predict the states of the system for more than 20 seconds
after the occlusion, thus permitting the successful capture of a tumbling satellite.

\bibliographystyle{IEEEtran}

\end{document}